\documentclass[12pt, a4paper]{article}

\usepackage[utf8]{inputenc}
\usepackage{amsmath, amssymb, amsfonts}
\usepackage{graphicx}
\usepackage[margin=1in]{geometry}
\usepackage{hyperref}
\usepackage{listings}
\usepackage{xcolor}
\usepackage{caption}
\usepackage{booktabs}
\usepackage{float}
\usepackage{url}
\usepackage{tikz}
\usetikzlibrary{arrows.meta,positioning}
\usepackage{enumitem}
\hypersetup{
    colorlinks=true,
    linkcolor=blue,
    filecolor=magenta,      
    urlcolor=cyan,
    pdftitle={PINGS Paper},
}

\definecolor{codegreen}{rgb}{0,0.6,0}
\definecolor{codegray}{rgb}{0.5,0.5,0.5}
\definecolor{codepurple}{rgb}{0.58,0,0.82}
\definecolor{backcolour}{rgb}{0.95,0.95,0.92}

\lstdefinestyle{mystyle}{
    backgroundcolor=\color{backcolour},   
    commentstyle=\color{codegreen},
    keywordstyle=\color{magenta},
    numberstyle=\tiny\color{codegray},
    stringstyle=\color{codepurple},
    basicstyle=\ttfamily\footnotesize,
    breakatwhitespace=false,         
    breaklines=true,                 
    captionpos=b,                    
    keepspaces=true,                 
    numbers=left,                    
    numbersep=5pt,                  
    showspaces=false,                
    showstringspaces=false,
    showtabs=false,                  
    tabsize=2
}
\begin{document}

\title{\textbf{PINGS: Physics-Informed Neural Network for Fast Generative Sampling}}


\author{
\begin{minipage}{\textwidth}\centering
\large
Achmad Ardani Prasha\thanks{Corresponding author: \texttt{41523010005@student.mercubuana.ac.id}},
Clavino Ourizqi Rachmadi,
\stepcounter{footnote} 
Muhamad Fauzan Ibnu Syahlan\thanks{Alumni, Universitas Mercu Buana, Jakarta, Indonesia},
Naufal Rahfi Anugerah,
Nanda Garin Raditya,
Putri Amelia,
Sabrina Laila Mutiara,
Hilman Syachr Ramadhan
\end{minipage}
\\[5pt]
\small Faculty of Computer Science, Universitas Mercu Buana, Jakarta, Indonesia
}

\maketitle

\begin{abstract}
We introduce \textbf{PINGS} (Physics-Informed Neural Network for Fast Generative Sampling),
a framework that \emph{amortizes} diffusion sampling by training a physics-informed network
to approximate reverse-time probability–flow dynamics, reducing sampling to a \emph{single}
forward pass (NFE=1).
As a \textbf{proof-of-concept}, we learn a direct map from a 3D standard normal to a non-Gaussian
Gaussian Mixture Model (GMM). PINGS preserves the target’s distributional structure
(multi-bandwidth kernel $\mathrm{MMD}^2=1.88\times10^{-2}$ with small errors on mean, covariance,
skewness, and excess kurtosis) and achieves constant-time generation:
$10^4$ draws in $16.54\pm0.56$\,ms on an RTX~3090, versus $468$–$843$\,ms for DPM–Solver (10/20)
and $960$\,ms for DDIM (50) under matched conditions.
We also sanity-check the PINN/automatic–differentiation pipeline on a damped harmonic oscillator,
obtaining MSEs down to $\mathcal{O}(10^{-5})$.
Compared to fast ODE solvers (iterative) and direct-map families (Flow/Rectified–Flow/Consistency),
PINGS frames generative sampling as a \emph{PINN-style residual} problem with endpoint anchoring,
yielding a white-box, differentiable map with NFE=1.
These proof-of-concept results position PINGS as a promising route to fast, function-based
generative sampling and suggest extensibility to scientific simulation (e.g., fast calorimetry).
\end{abstract}

\section{Introduction}

\subsection{The Rise and Challenge of Diffusion Models}

Generative modeling has witnessed a significant leap forward with the advent of diffusion models, also known as score-based generative models \cite{goodfellow2014gan}. These models have demonstrated an unparalleled ability to generate high-fidelity samples across a multitude of domains, including photorealistic image synthesis, audio generation, and complex scientific data modeling \cite{ho2020ddpm}. The core principle of these models involves a two-stage process: a fixed "forward" process that systematically corrupts data with noise until it resembles a simple prior distribution (typically Gaussian), and a learned "reverse" process that iteratively denoises samples from the prior to generate new data \cite{ho2020ddpm}.
Despite their success, the practical deployment of diffusion models is often hampered by a significant computational bottleneck: the sampling process is inherently slow and resource-intensive. The reverse denoising process is typically modeled as a long Markov chain or as the discretization of a stochastic differential equation (SDE), requiring hundreds or even thousands of sequential evaluations of a large neural network to produce a single sample \cite{lu2022dpm}. This high inference cost presents a major obstacle for applications requiring real-time generation or large-scale data simulation.

\subsection{Accelerating Sampling via the Probability Flow ODE}

A key theoretical development in addressing this challenge was the realization that the stochastic reverse process has a deterministic counterpart governed by an Ordinary Differential Equation (ODE), often referred to as the probability flow ODE \cite{song2021sde}. Solving this ODE from its terminal time to its initial time generates samples from the same distribution as the SDE but with reduced stochasticity, which can lead to faster and more stable sampling \cite{song2021sde}. This insight has spurred the development of specialized high-order ODE solvers tailored for diffusion models, such as the DPM-Solver \cite{lu2022dpm}. These methods can drastically reduce the number of required function evaluations from thousands to as few as 10-20, representing a significant acceleration \cite{lu2022dpm}. However, even these advanced techniques are fundamentally iterative, requiring multiple sequential calls to the underlying neural network.

\subsection{An Alternative: Physics-Informed Neural Networks (PINNs)}

Concurrently, a distinct paradigm for solving differential equations has emerged from the scientific machine learning community: Physics-Informed Neural Networks (PINNs) \cite{raissi2019pinn}. PINNs are neural networks trained to directly approximate the solution of a differential equation by incorporating the equation itself into the network's loss function. The loss penalizes the network's output for deviating from the governing physical law (the differential equation residual) and for failing to meet the specified boundary or initial conditions \cite{raissi2019pinn}. Enabled by automatic differentiation, PINNs can learn continuous, mesh-free solutions to complex ODEs and PDEs, positioning them as powerful function approximators for physical systems \cite{cuomo2022scientific}.

\subsection{Our Contribution: The PINGS Framework}
This paper introduces PINGS (Physics-Informed Neural network for Fast Generative Sampling), a novel framework that synthesizes diffusion-based generative modeling with Physics-Informed Neural Networks. Unlike prior approaches that rely on iterative solvers or that learn only components of the reverse process, PINGS directly learns the entire solution to the reverse-time probability flow ODE \cite{lipman2022flow}. Once trained, PINGS acts as a direct generative mapping, reducing sampling to a single forward pass from a random noise vector to a high-fidelity data sample.
This approach fundamentally changes the generative modeling workflow. The computationally intensive task of iteratively solving the ODE is amortized into a one-time, upfront training cost for the PINN. Once trained, PINGS itself becomes the generative model. Sampling is reduced to a single forward evaluation, in stark contrast to conventional methods that require iterative calls to a score network or expensive Monte Carlo integration \cite{salimans2022progressive}.
In this work, we demonstrate the efficacy of PINGS on the task of fast generative sampling of non-Gaussian 3D densities, showing that it can accurately reproduce statistical properties such as mean, covariance, skewness, and kurtosis while achieving orders-of-magnitude speedup compared to iterative diffusion samplers. This paper details the theoretical underpinnings of PINGS, validates its implementation on a canonical physics problem, and demonstrates its accuracy and efficiency for complex 3D density estimation tasks.

\subsection{Related Work and Positioning}
\label{sec:related}
High–order ordinary differential equation (ODE) samplers such as DPM–Solver~\cite{lu2022dpm} accelerate probability–flow integration but remain iterative (large NFE). Flow Matching (FM)~\cite{lipman2022flow},Consistency Models (CM)~\cite{song2023consistency}, and Rectified Flow (RF)~\cite{liu2022flowstraight} train vector fields or consistency objectives so that sampling is non-iterative. 
PINGS (Physics-Informed Neural Network for Fast Generative Sampling) casts generation as a physics-informed neural network (PINN) residual over $t\in(0,1)$, amortizing the probability–flow ODE into a single function $g_\theta$ anchored by: (i) identity at $t{=}1$; (ii) distribution match at $t{=}0$ via MMD$^2$ and moment alignment; and (iii) an interior residual that is either score-informed or score-free. Sampling achieves NFE$=1$ via analytic evaluation. Unlike FM/RF/CM, PINGS inherits PINN traits—exact derivatives via automatic differentiation (AD)—yielding a white-box generator with tractable sensitivities alongside fast sampling.

\section{Theoretical Framework}

\subsection{Score-Based Generative Modeling via Diffusion Processes}

\subsubsection{Forward Process (Data to Noise)}

The foundation of a diffusion model is the forward process, which describes how a data sample $\mathbf{x}_0$ from a complex data distribution $p_{\text{data}}(\mathbf{x})$ is gradually transformed into a sample $\mathbf{x}_T$ from a simple, tractable prior distribution $p_{\text{prior}}(\mathbf{x})$, typically a standard normal distribution $\mathcal{N}(0,I)$. In the continuous-time formulation, this transformation is governed by an Itô Stochastic Differential Equation (SDE) of the form \cite{song2021sde}:
\begin{equation}
    d\mathbf{x} = \mathbf{f}(\mathbf{x}, t)dt + g(t)d\mathbf{w}
\end{equation}
Here, $\mathbf{x}(t)$ represents the state of the sample at time $t \in [0, T]$, $\mathbf{f}(\mathbf{x},t)$ is a drift function that controls the deterministic evolution of the sample, $g(t)$ is a diffusion coefficient that scales the magnitude of the injected noise, and $d\mathbf{w}$ is an infinitesimal step of a standard Wiener process (Brownian motion). The choice of $\mathbf{f}$ and $g$ defines the specific diffusion process, with common choices leading to Variance Preserving (VP) or Variance Exploding (VE) SDEs \cite{song2021sde}.

\subsubsection{Reverse Process (Noise to Data)}

A remarkable result, established by Anderson \cite{anderson1982reverse}, is that under mild regularity conditions, the forward SDE process is reversible in time. The corresponding reverse-time SDE, which evolves from the prior distribution at time
$T$ back to the data distribution at time 0, is given by \cite{anderson1982reverse}:
\begin{equation}
    d\mathbf{x} = [\mathbf{f}(\mathbf{x}, t) - g(t)^2\nabla_{\mathbf{x}}\log p_t(\mathbf{x})]dt + g(t)d\bar{\mathbf{w}}
\end{equation}
where $d\bar{\mathbf{w}}$ is a standard Wiener process running backward in time, and $p_t(\mathbf{x})$ is the marginal probability density of the perturbed data $\mathbf{x}(t)$.

\subsubsection{The Score Function}

The critical term in the reverse SDE is $\nabla_{\mathbf{x}}\log p_t(\mathbf{x})$, known as the score function of the distribution $p_t$ \cite{song2021sde}. This vector field points in the direction of the steepest ascent of the log-probability density. The term
$-g(t)^2\nabla_{\mathbf{x}}\log p_t(\mathbf{x})$ acts as a corrective drift, guiding the samples from low-probability regions toward high-probability regions, effectively "denoising" them and reversing the diffusion process \cite{song2021sde}. Since
$p_t(\mathbf{x})$ is generally intractable, the core task of score-based models is to train a neural network, $\mathbf{s}_\theta(\mathbf{x}, t)$, to approximate the score function.

\subsubsection{The Probability Flow ODE}

For every SDE, there exists a corresponding deterministic Ordinary Differential Equation (ODE) whose trajectories share the same marginal probability densities $p_t(\mathbf{x})$ over time \cite{song2021sde}. This probability flow ODE is given by:
\begin{equation}
    \frac{d\mathbf{x}}{dt} = \mathbf{f}(\mathbf{x}, t) - \frac{1}{2}g(t)^2\nabla_{\mathbf{x}}\log p_t(\mathbf{x})
\end{equation}
Solving this ODE backward in time from $t=T$ to $t=0$ provides a deterministic mapping from a sample $\mathbf{x}_T \sim p_{\text{prior}}$ to a sample $\mathbf{x}_0 \sim p_{\text{data}}$. This ODE forms the physical law that our PINN will be trained to solve, eliminating the need for a stochastic solver and providing a direct path for generation.

\subsection{Physics-Informed Neural Networks (PINNs)}

\subsubsection{Core Concept}
Physics-Informed Neural Networks (PINNs) are a class of neural networks specifically designed to solve differential equations by embedding the physical laws they represent directly into the learning process \cite{raissi2019pinn}. The fundamental idea is to use a neural network,
$u_\theta(\mathbf{x})$, as a universal function approximator for the solution of a differential equation, which can be expressed in its residual form as $\mathcal{N}[u(\mathbf{x})] = 0$ for $\mathbf{x}$ in a domain $\Omega$ \cite{raissi2019pinn}.

\subsubsection{The Physics-Informed Loss Function}
The training of a PINN is guided by a composite loss function that enforces both the governing equation and its associated boundary or initial conditions \cite{raissi2019pinn}. The total loss, $\mathcal{L}_{\text{total}}(\theta)$, is typically a weighted sum of two components:
\begin{equation}
    \mathcal{L}_{\text{total}}(\theta) = w_r\mathcal{L}_{\text{residual}} + w_b\mathcal{L}_{\text{boundary}}
\end{equation}
\textbf{Residual Loss ($\mathcal{L}_{\text{residual}}$):} This term measures how well the network's output satisfies the differential equation. It is formulated as the mean squared error of the PDE residual, evaluated over a set of $N_r$ "collocation points" randomly sampled from the domain $\Omega$.
\begin{equation}
    \mathcal{L}_{\text{residual}} = \frac{1}{N_r} \sum_{i=1}^{N_r} \|\mathcal{N}[u_\theta(\mathbf{x}_r^i)]\|^2
\end{equation}
This loss term forces the network to learn a function that is consistent with the underlying physics \cite{raissi2019pinn}. \\
\textbf{Boundary Loss ($\mathcal{L}_{\text{boundary}}$):} This term ensures that the learned solution respects the problem's constraints. It is the mean squared error between the network's prediction and the known values at the $N_b$ boundary points.
\begin{equation}
    \mathcal{L}_{\text{boundary}} = \frac{1}{N_b} \sum_{i=1}^{N_b} \|u_\theta(\mathbf{x}_b^i) - u_{\text{true}}(\mathbf{x}_b^i)\|^2
\end{equation}
This includes initial conditions, Dirichlet boundary conditions, and Neumann boundary conditions \cite{raissi2019pinn}.

\subsubsection{Automatic Differentiation}
The enabling technology for PINNs is automatic differentiation (AD), a cornerstone of modern deep learning frameworks like PyTorch \cite{raissi2019pinn}. AD allows for the exact computation of the derivatives of the neural network's output with respect to its inputs (e.g.,
$\frac{\partial u_\theta}{\partial x}, \frac{\partial^2 u_\theta}{\partial x^2}$). These derivatives are essential for constructing the residual loss term $\mathcal{L}_{\text{residual}}$ without resorting to numerical approximations, thereby allowing the physics to be encoded with high fidelity \cite{raissi2019pinn}.

\subsubsection{Advantages and Challenges}
As surveyed by Cuomo et al. \cite{cuomo2022scientific}, PINNs offer several compelling advantages over traditional numerical solvers. They are mesh-free, making them well-suited for problems with complex geometries or high dimensions \cite{somvanshi2024not}. They can also seamlessly solve inverse problems (e.g., inferring PDE parameters from data) within the same framework \cite{somvanshi2024not}. However, training PINNs is not without challenges. The optimization landscape can be highly non-convex, and the training process can suffer from issues like vanishing or exploding gradients, particularly for stiff or multi-scale problems \cite{wang2021understanding}. Successfully training a PINN often requires careful tuning of the network architecture, optimizer, and the relative weighting of the loss components.

\section{Methodology: The PINGS Model}
\label{sec:method}
\subsection{Governing Equation Formulation}
Let $\mathbf{z}\sim\mathcal{N}(\mathbf{0},\mathbf{I}_3)$ and $t\in[0,1]$. We learn a continuous flow
$\mathbf{x}(t)=g_\theta(t,\mathbf{z})$ with boundary conditions $g_\theta(1,\mathbf{z})=\mathbf{z}$ and
$g_\theta(0,\cdot)\sim p_{\mathrm{data}}$. Instead of integrating a probability–flow ODE with an explicit
VP–SDE schedule, we directly penalize the residual
\begin{equation}
  \label{eq:resid}
  \mathcal{R}(t,\mathbf{z})
  \;=\;
  \partial_t g_\theta(t,\mathbf{z})
  \;-\;
  \alpha(t)\, s\!\big(g_\theta(t,\mathbf{z})\big),
  \qquad
  \alpha(t)=\alpha_{\mathrm{scale}}\,(1-t)^{\alpha_{\mathrm{power}}},
\end{equation}
where $s(\cdot)$ is a score function. We consider two training modes:
(i) \textbf{score-informed} — when an explicit score is available (our synthetic Gaussian Mixture Model (GMM)),
$s(\cdot)$ is set to the closed‐form mixture score; and
(ii) \textbf{score-free} — when no explicit score is available, we set $s(\cdot)\equiv 0$,
so $\mathcal{R}(t,\mathbf{z})=\partial_t g_\theta(t,\mathbf{z})$ and rely on boundary and
distribution matching to anchor the endpoints. The time derivative $\partial_t g_\theta$
is obtained via automatic differentiation.

\paragraph{Interpretation of the residual.}
Eq.~\eqref{eq:resid} encourages the trajectory $t\mapsto g_\theta(t,\mathbf z)$ to follow
probability–flow dynamics \emph{without} explicitly integrating an ODE. The term
$\partial_t g_\theta(t,\mathbf z)$ is the instantaneous velocity (via auto–diff), while
$\alpha(t)\,s(g_\theta(t,\mathbf z))$ is a target velocity that steers samples toward
higher log–density regions, scaled by $\alpha(t)$. With
$\alpha(t)=\alpha_{\mathrm{scale}}(1-t)^{\alpha_{\mathrm{power}}}$, we have
$\alpha(1)=0$ so the flow can satisfy the identity at $t{=}1$, and
$\alpha(0)=\alpha_{\mathrm{scale}}$ so score guidance is strongest near $t{=}0$ where
outputs must match $p_{\mathrm{data}}$. In the score–free variant ($s\equiv 0$),
interior dynamics are regularized by $\mathcal{L}_{\mathrm{phys}}$, while endpoints are
enforced by $\mathcal{L}_{\mathrm{bc}}$ and $\mathcal{L}_{\mathrm{mmd}}{+}\mathcal{L}_{\mathrm{mom}}$.

\begin{samepage}
\paragraph{Closed-form score for the 3D Gaussian mixture.}
For $p(\mathbf x)=\sum_{k=1}^{K}\pi_k\,\mathcal N(\mathbf x;\boldsymbol\mu_k,\Sigma_k)$ with diagonal
$\Sigma_k=\mathrm{diag}(\boldsymbol\sigma_k^2)$, the score is
\begin{equation}
  \label{eq:gmm-score}
  \begin{aligned}
    s(\mathbf x)
      &= \nabla_{\mathbf x}\log p(\mathbf x)
       = \sum_{k=1}^K r_k(\mathbf x)\,\Sigma_k^{-1}\!\big(\boldsymbol\mu_k-\mathbf x\big),\\[-2pt]
    \text{where}\quad
    r_k(\mathbf x)
      &= \frac{\pi_k\,\mathcal N(\mathbf x;\boldsymbol\mu_k,\Sigma_k)}
              {\sum_{j=1}^{K}\pi_j\,\mathcal N(\mathbf x;\boldsymbol\mu_j,\Sigma_j)}.
  \end{aligned}
\end{equation}
Here $r_k(\mathbf x)$ are the responsibilities (posterior component probabilities).
\end{samepage}

\subsection{Network Architecture}
To model the non–linear mapping from the 4D input $(t,\mathbf{z})$ to the 3D output $\mathbf{x}$,
we use a fully–connected network $g_\theta$ with an input of size $4$ for $(t,z_1,z_2,z_3)$, six
hidden layers of width $128$ using $\tanh$ activations and Xavier initialization (bias~$=0$),
and a linear output of size $3$ yielding $\mathbf{x}=(x_1,x_2,x_3)$. In shorthand, the
architecture is $4 \rightarrow 128 \rightarrow 128 \rightarrow 128 \rightarrow 128 \rightarrow 128
\rightarrow 128 \rightarrow 3$. This design yields smooth derivatives for $\partial_t g_\theta$
(via auto–diff) while providing sufficient capacity to approximate the target flow.

\begin{figure}[H]
  \centering
  \begin{tikzpicture}[>=Latex, node distance=8mm, font=\small]
    \tikzstyle{io}    =[draw, rounded corners=2pt, minimum width=4.2cm, minimum height=7mm, align=center, fill=blue!8]
    \tikzstyle{layer} =[draw, rounded corners=2pt, minimum width=4.2cm, minimum height=7mm, align=center, fill=gray!10]
    \tikzstyle{arrow} =[->, line width=0.6pt]

    \node[io]    (in)  {Input (4): $t,\, z_1,\, z_2,\, z_3$};
    \node[layer] (h1)  [below=of in]  {Dense 128, Tanh};
    \node[layer] (h2)  [below=of h1]  {Dense 128, Tanh};
    \node[layer] (h3)  [below=of h2]  {Dense 128, Tanh};
    \node[layer] (h4)  [below=of h3]  {Dense 128, Tanh};
    \node[layer] (h5)  [below=of h4]  {Dense 128, Tanh};
    \node[layer] (h6)  [below=of h5]  {Dense 128, Tanh};
    \node[io]    (out) [below=of h6]  {Output (3): $x_1,\, x_2,\, x_3$ (Linear)};

    \draw[arrow] (in) -- (h1);
    \draw[arrow] (h1) -- (h2);
    \draw[arrow] (h2) -- (h3);
    \draw[arrow] (h3) -- (h4);
    \draw[arrow] (h4) -- (h5);
    \draw[arrow] (h5) -- (h6);
    \draw[arrow] (h6) -- (out);
  \end{tikzpicture}
  \caption{Architecture of \textbf{PINGS}: a Multilayer Perceptron (MLP) mapping $(t,\mathbf z)\!\in\!\mathbb R^4$
  to $\mathbf x\!\in\!\mathbb R^3$ with six 128–unit $\tanh$ layers and a linear output.}
  \label{fig:pings_arch}
\end{figure}
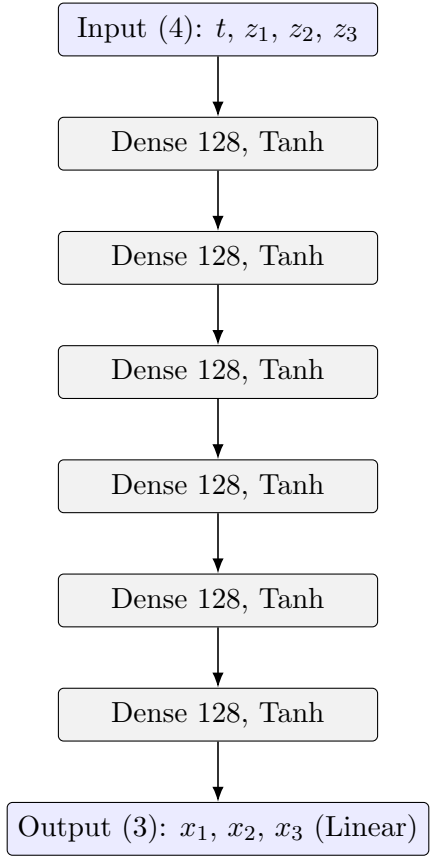

\subsection{Training Objectives}
The total objective is
\begin{equation}
  \mathcal{L}(\theta)=
  \lambda_{\mathrm{bc}}\mathcal{L}_{\mathrm{bc}}
  +\lambda_{\mathrm{mmd}}\mathcal{L}_{\mathrm{mmd}}
  +\lambda_{\mathrm{mom}}\mathcal{L}_{\mathrm{mom}}
  +\lambda_{\mathrm{phys}}\mathcal{L}_{\mathrm{phys}},
\end{equation}
with $(\lambda_{\mathrm{bc}},\lambda_{\mathrm{mmd}},\lambda_{\mathrm{mom}},\lambda_{\mathrm{phys}})
=(1.0,\,2.0,\,0.1,\,0.5)$.

\paragraph{Boundary identity at $t{=}1$.}
\begin{equation}
  \mathcal{L}_{\mathrm{bc}}
  = \mathbb E_{\mathbf z}\,\|g_\theta(1,\mathbf z)-\mathbf z\|_2^2.
\end{equation}

\paragraph{Distribution matching at $t{=}0$.}
We match $g_\theta(0,\mathbf Z)$ to $\mathbf X\!\sim p_{\mathrm{data}}$ using two criteria.
First, the unbiased squared Maximum Mean Discrepancy (MMD) with a multi–Gaussian kernel bank,
\begin{equation}
  \mathrm{MMD}^2(\mathcal A,\mathcal B)
  = \frac{1}{m(m-1)}\!\sum_{i\neq j} k(\mathbf a_i,\mathbf a_j)
  + \frac{1}{n(n-1)}\!\sum_{i\neq j} k(\mathbf b_i,\mathbf b_j)
  - \frac{2}{mn}\!\sum_{i,j} k(\mathbf a_i,\mathbf b_j),
\end{equation}
where $\mathcal A=\{g_\theta(0,\mathbf z_i)\}_{i=1}^m$, $\mathcal B=\{\mathbf x_j\}_{j=1}^n$, and
$k(\mathbf u,\mathbf v)=\sum_{\sigma\in\mathcal S}\exp\!\big(-\|\mathbf u-\mathbf v\|_2^2/(2\sigma^2)\big)$
with $\mathcal S=\{0.1,0.2,0.5,1.0,2.0\}$. Second, a moment loss aligning first and second moments,
\begin{equation}
  \mathcal{L}_{\mathrm{mom}}
  = \|\mu_\theta-\mu\|_2^2 + \|\Sigma_\theta-\Sigma\|_F^2,
\end{equation}
where $(\mu_\theta,\Sigma_\theta)$ are empirical mean and covariance of $g_\theta(0,\mathbf Z)$ and
$(\mu,\Sigma)$ are those of $\mathbf X$; $\|\cdot\|_F$ is the Frobenius norm. We also \emph{report}
(but do not optimize) skewness and excess kurtosis to assess higher–order moment fidelity.

\paragraph{Physics-like residual on $t\in(0,1)$.}
\begin{equation}
  \mathcal{L}_{\mathrm{phys}}
  = \mathbb E_{t\sim\mathcal U(0,1),\,\mathbf z}\,\|\mathcal R(t,\mathbf z)\|_2^2,
  \quad\text{with }\mathcal R \text{ given in Eq.~(\ref{eq:resid})}.
\end{equation}
This ties the entire path between $t{=}1$ and $t{=}0$, preventing shortcuts that satisfy only the endpoints.

\subsection{Implementation and Training}
At each iteration, we draw independent samples $\mathbf z\sim\mathcal N(\mathbf 0,\mathbf I_3)$ and
$t\sim\mathcal U(0,1)$ to populate the training batches. The boundary loss $\mathcal{L}_{\mathrm{bc}}$
is evaluated at $t=1$ to enforce the identity mapping $g_\theta(1,\mathbf z)=\mathbf z$, while the
distribution–matching terms $\mathcal{L}_{\mathrm{mmd}}$ and $\mathcal{L}_{\mathrm{mom}}$ are evaluated at
$t=0$ to align $g_\theta(0,\mathbf Z)$ with $p_{\mathrm{data}}$. The physics–like residual
$\mathcal{L}_{\mathrm{phys}}$ is applied at interior points $t\in(0,1)$, tying the trajectory between the
two endpoints so that the model does not satisfy the constraints only at $t=0$ and $t=1$ but also along
the path. The function $g_\theta$ is implemented as a fully connected multilayer perceptron with six hidden layers
of width 128 using $\tanh$ activations. All weights are initialized with Xavier initialization and biases
are set to zero. This choice yields smooth derivatives for $\partial_t g_\theta$ via automatic differentiation
while providing enough capacity to approximate the target flow.

\vspace{1em} 
\noindent We train with Adam (learning rate $10^{-3}$) and apply an exponential decay with factor $\gamma=0.999$
every 1000 epochs. Unless otherwise stated, we use a batch size of 2048, early stopping with a patience of
3000 steps, and a fixed random seed (42) for reproducibility. The schedule parameters are set to
$\alpha_{\mathrm{scale}}=1.0$ and $\alpha_{\mathrm{power}}=1.0$, which makes the guidance strongest near
$t=0$ and vanish at $t=1$.

\vspace{1em}
\noindent At inference time, sampling $N$ points reduces to a single batched forward pass $g_\theta(0,\mathbf Z)$,
where $\mathbf Z\sim\mathcal N(\mathbf 0,\mathbf I_3)$. To obtain reliable wall–clock measurements, we
enable CUDA Deep Neural Network library (cuDNN) benchmarking and Automatic Mixed Precision (AMP) during inference, and we wrap each timed forward pass
with \texttt{torch.cuda.synchronize()} before and after execution. For $N=10{,}000$ samples, we report the
mean $\pm$ standard deviation over five runs.

\section{Tooling Validation on a Canonical ODE (Damped Harmonic Oscillator (DHO))}
\label{sec:dho}

\subsection{Problem and Setup}
This subsection is a tooling sanity check for the
PINN/automatic-differentiation (AD) pipeline used by \textbf{PINGS}. We verify (i) enforcement of
value and derivative initial conditions via AD, (ii) stable optimization of the residual on
randomly resampled collocation points, and (iii) numerical accuracy of first/second derivatives.
It does \emph{not} evaluate distribution-matching objectives (MMD/moment) or score-informed
residuals and is therefore \emph{not} an apple-to-apple comparison with the 3D generative task;
its sole purpose is to certify numerical correctness before Sec.~\ref{sec:primary_3d}. We validate our PINN setup on a well-studied ODE with a closed-form solution: the damped harmonic oscillator. The governing ODE and domain/Initial Conditions (ICs) are
\begin{equation}
\label{eq:dho_ode}
\begin{alignedat}{2}
\frac{d^2 x}{dz^2} + 2\,\xi\,\frac{dx}{dz} + x \;&=\; 0,\\
x(0) \;&=\; 0.7, \qquad \left.\frac{dx}{dz}\right|_{z=0} \;=\; 1.2,\\
\xi \;&\in\; [0.1,0.4], \qquad z \in [0,20].
\end{alignedat}
\end{equation}

\noindent For the underdamped regime $\xi<1$, the analytical (ground-truth) solution used for evaluation is
\begin{equation}
  \label{eq:dho_analytical}
  \begin{aligned}
    x_{\mathrm{ana}}(z;\xi)
      &= e^{-\xi z}\!\left(C_{1}\cos(\omega_{d} z)+C_{2}\sin(\omega_{d} z)\right),\\
    \omega_{d} &= \sqrt{1-\xi^{2}}, \qquad
    C_{1} = 0.7, \qquad
    C_{2} = \frac{1.2+\xi C_{1}}{\omega_{d}}.
  \end{aligned}
\end{equation}

\noindent To avoid numerical issues when $\xi$ approaches $1$, we implement
\begin{equation}
  \label{eq:dho_safe_omega}
  \omega_d \;=\; \sqrt{\max(1-\xi^2,\,10^{-12})}.
\end{equation}

\subsection{PINN Formulation}
We learn a scalar surrogate $x_\theta(z,\xi)$ with a fully connected network (FCNN) taking $(z,\xi)\in\mathbb{R}^2$ and outputting $x\in\mathbb{R}$. The training objective is the sum of three terms,
\begin{equation}
  \label{eq:dho_loss_total}
  \mathcal{L}_{\mathrm{osc}} \;=\; \mathcal{L}_{\mathrm{ic,val}} \;+\; \mathcal{L}_{\mathrm{ic,der}} \;+\; \mathcal{L}_{\mathrm{res}},
\end{equation}
with
\begin{align}
  \label{eq:dho_loss_ic_val}
  \mathcal{L}_{\mathrm{ic,val}}
  &= \frac{1}{N_{\mathrm{ic}}}\sum_{i=1}^{N_{\mathrm{ic}}}\!\Big(x_\theta(0,\xi_i)-0.7\Big)^2, \\[4pt]
  \label{eq:dho_loss_ic_der}
  \mathcal{L}_{\mathrm{ic,der}}
  &= \frac{1}{N_{\mathrm{ic}}}\sum_{i=1}^{N_{\mathrm{ic}}}\!
     \left(\left.\frac{d x_\theta}{dz}\right|_{(0,\xi_i)} - 1.2 \right)^{\!2}, \\[4pt]
  \label{eq:dho_loss_res}
  \mathcal{L}_{\mathrm{res}}
  &= \frac{1}{N_{r}}\sum_{j=1}^{N_{r}}
     \left(\left.\frac{d^2 x_\theta}{dz^2}\right|_{(z_j,\xi_j)}
     + 2\,\xi_j \left.\frac{d x_\theta}{dz}\right|_{(z_j,\xi_j)}
     + x_\theta(z_j,\xi_j)\right)^{\!2}.
\end{align}
All derivatives are computed exactly via automatic differentiation (nested calls to \path{torch.autograd.grad}).
We use an FCNN with layers $[2,32,32,32,32,1]$, $\tanh$ activations, and Xavier initialization (zero bias). At each epoch we resample training points:
(i) $N_{\mathrm{ic}}{=}100$ IC points with $z{=}0$, $\xi\sim\mathcal{U}[0.1,0.4]$;
(ii) $N_r{=}2000$ collocation points with $z\sim\mathcal{U}[0,20]$, $\xi\sim\mathcal{U}[0.1,0.4]$.
We train for up to $20{,}000$ epochs with Adam ($\mathrm{lr}{=}10^{-3}$), an exponential scheduler with factor $\gamma{=}0.99$ applied every $1000$ epochs, and early stopping (patience $1500$). We fix the random seed to $42$.

\subsection{Results and Analysis}
\begin{figure}[H]
  \centering
  \includegraphics[width=\textwidth]{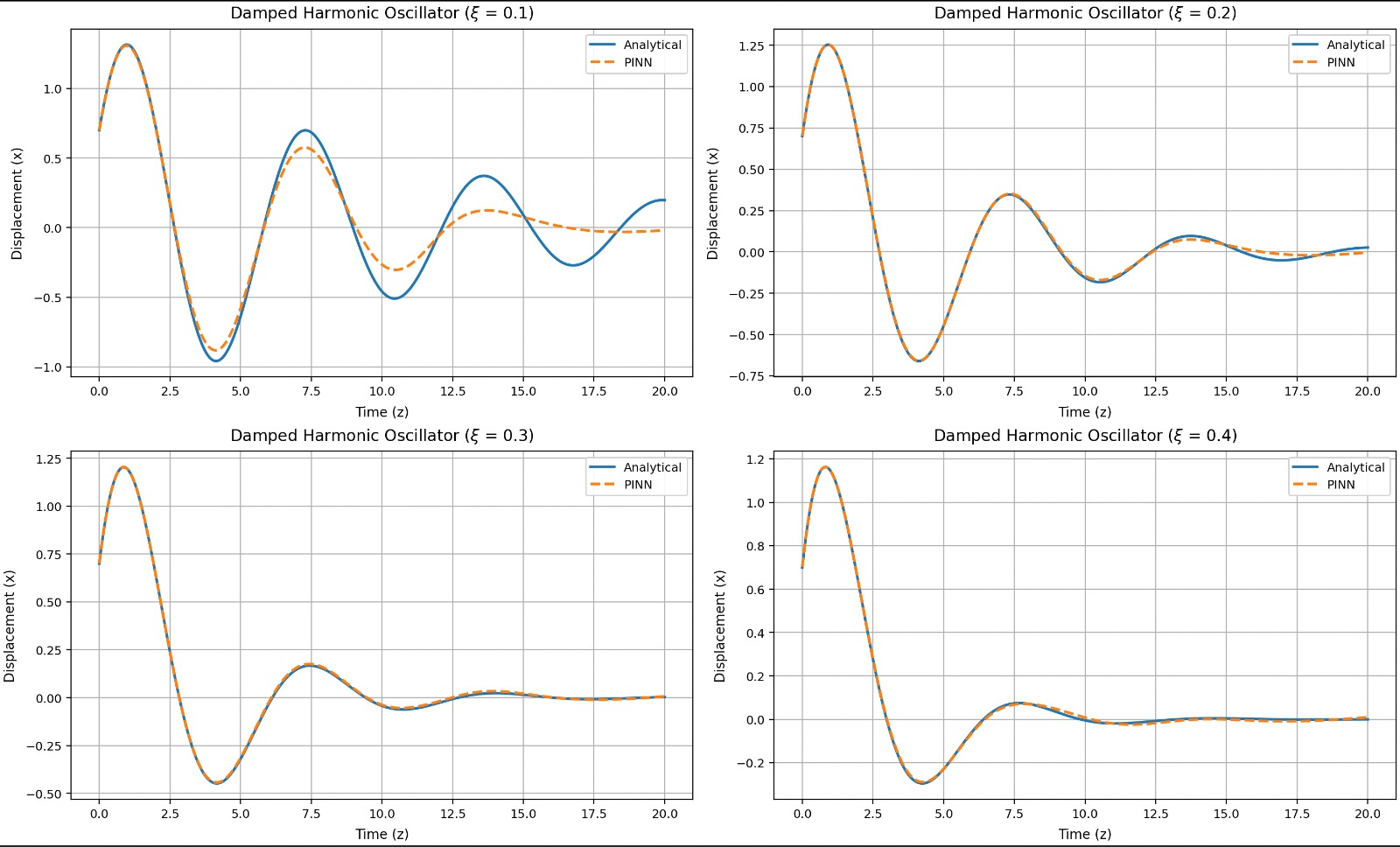}
  \caption{Comparison of PINN predictions and analytical solutions for the damped harmonic oscillator at $\xi\in\{0.1,0.2,0.3,0.4\}$.}
  \label{fig:oscillator}
\end{figure}

Figure~\ref{fig:oscillator} compares the PINN prediction against the analytical solution \eqref{eq:dho_analytical} at four damping ratios. The model captures both frequency and amplitude decay accurately across the range.

\noindent For a quantitative assessment we compute the mean-squared error (MSE) between $x_\theta$ and $x_{\mathrm{ana}}$ on a fine grid ($z\in[0,20]$, $5000$ points) for each $\xi$. The results are:

\begin{table}[H]
  \centering
  \caption{MSE of PINN vs.\ analytical solution over $z\in[0,20]$ for each damping ratio.}
  \label{tab:oscillator_mse}
  \begin{tabular}{@{}ll@{}}
    \toprule
    \textbf{Damping Ratio ($\xi$)} & \textbf{MSE} \\
    \midrule
    $0.1$ & $1.77\times 10^{-2}$ \\
    $0.2$ & $2.39\times 10^{-4}$ \\
    $0.3$ & $3.56\times 10^{-5}$ \\
    $0.4$ & $4.51\times 10^{-5}$ \\
    \bottomrule
  \end{tabular}
\end{table}

\noindent Errors are $\mathcal{O}(10^{-5})$ for
$\xi\!\in\!\{0.3,0.4\}$ and $2.39\times 10^{-4}$ at $\xi\!=\!0.2$, with a larger yet acceptable
$1.77\times 10^{-2}$ at $\xi\!=\!0.1$ (lightly damped, long-lived oscillations). These results
confirm that our AD-based residuals, IC handling, and optimization are correct on a closed-form
ODE. While this check does not probe the generative components (MMD/moments, score-informed
residuals, higher dimension), it validates the numerical backbone used in the 3D experiments of
Sec.~\ref{sec:primary_3d}.

\section{Primary Application: Fast Generative Sampling of Non-Gaussian 3D Densities}
\label{sec:primary_3d}

\subsection{Experimental Setup}
Having validated the PINN tooling on the DHO, we apply PINGS to learn a direct mapping
from a 3D standard normal prior to a non-Gaussian target. We operate in the
\emph{score-informed} mode (see Sec.~\ref{sec:method}): the interior residual uses the
closed-form score of the target mixture, while the boundary identity at $t{=}1$ and
distribution matching at $t{=}0$ (MMD$^2$ + moment alignment) anchor the endpoints.
Concretely, the training objective is
\begin{equation}
  \mathcal L(\theta) \;=\; \lambda_{\text{bc}}\mathcal L_{\text{bc}}
  + \lambda_{\text{mmd}}\mathcal L_{\text{mmd}}
  + \lambda_{\text{mom}}\mathcal L_{\text{mom}}
  + \lambda_{\text{phys}}\mathcal L_{\text{phys}},
\end{equation}
with weights specified in Sec.~\ref{sec:method}.
Gradients \emph{with respect to} time $t$ for the residual term are obtained by automatic
differentiation, so the entire flow $g_\theta(t,\mathbf z)$ is learned as a single continuous function.

\subsubsection{Target Distribution}
The target density is a three-component diagonal-covariance Gaussian mixture:
\begin{subequations}
\label{eq:target_gmm_all}
\begin{align}
p_{\text{target}}(\mathbf x)
  &= \sum_{k=1}^{3}\pi_k \,\mathcal N\!\big(\mathbf x;\,\boldsymbol\mu_k,\,\Sigma_k\big), \label{eq:target_gmm}\\
\boldsymbol\pi &= [\,0.5,\,0.3,\,0.2\,], \qquad \Sigma_k=\mathrm{diag}(\boldsymbol\sigma_k^2), \label{eq:target_weights}\\
\boldsymbol\mu_1&=[\,2.5,\,0.0,\,-1.5\,],\;
\boldsymbol\mu_2=[\,-2.0,\,2.0,\,1.0\,],\;
\boldsymbol\mu_3=[\,0.0,\,-2.5,\,2.0\,], \label{eq:target_means}\\
\boldsymbol\sigma_1&=[\,0.60,\,0.50,\,0.70\,],\;
\boldsymbol\sigma_2=[\,0.45,\,0.65,\,0.40\,],\;
\boldsymbol\sigma_3=[\,0.55,\,0.40,\,0.60\,]. \label{eq:target_stds}
\end{align}
\end{subequations}
The residual uses the analytical mixture score $s(\mathbf x)$ defined in Eq.~(\ref{eq:gmm-score}).

\subsubsection{Training Details}
Unless stated otherwise: $20{,}000$ epochs; batch size $2048$ (prior and target);
Adam ($10^{-3}$) with an exponential scheduler $\gamma{=}0.999$ applied each $1000$ epochs;
early stopping (patience $3000$); seed $42$. Hardware: NVIDIA RTX~3090. We enable
cuDNN benchmarking and mixed precision (AMP) for inference timing. (MLP: input $(t,\mathbf z)\!\in\!\mathbb R^4$, six hidden layers $\times\,128$ with \texttt{tanh};
weights Xavier, zero biases; see Sec.~\ref{sec:method}.)

\subsection{Qualitative Results}
\begin{figure}[H]
  \centering
  \includegraphics[width=0.78\textwidth]{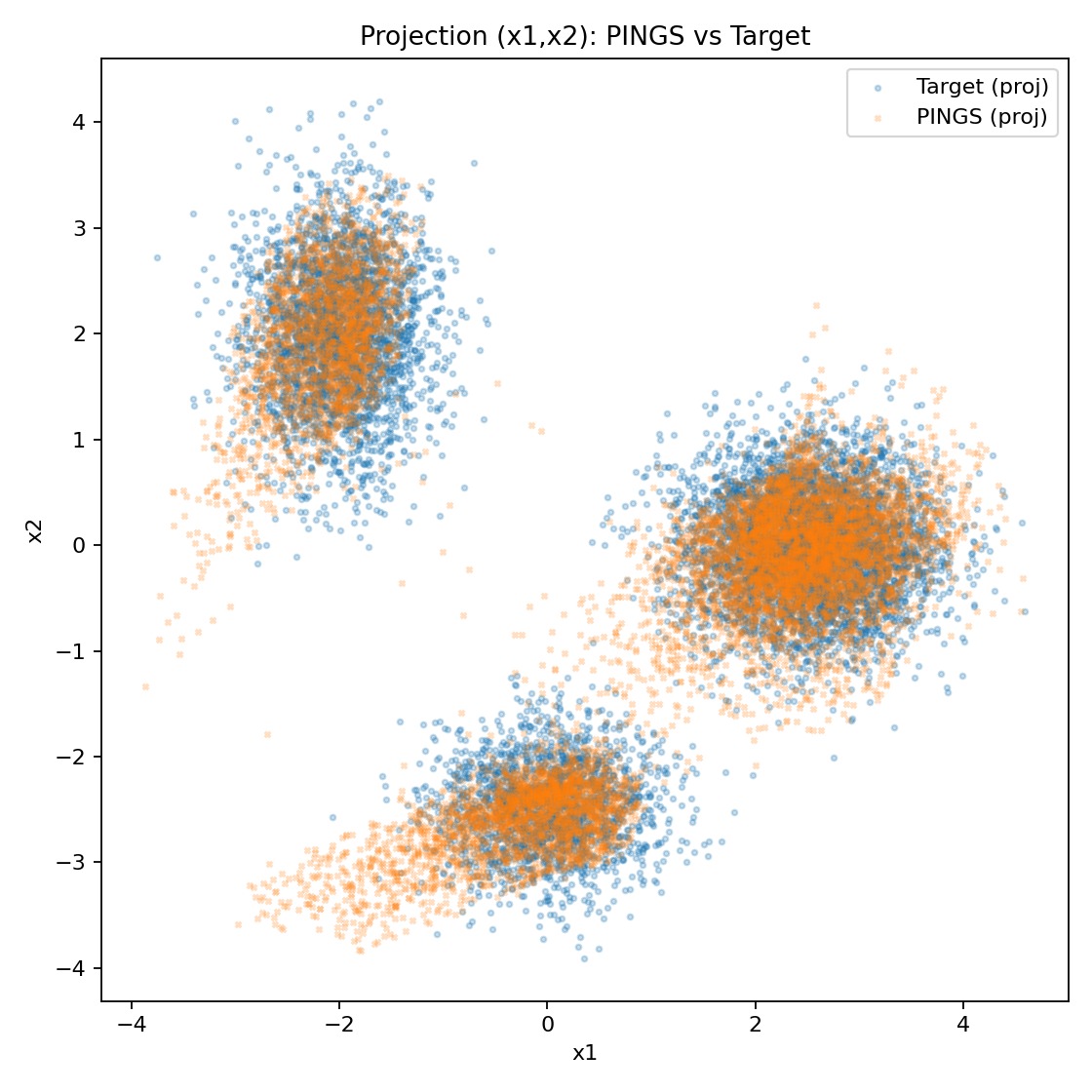}
  \caption{Projection $(x_1,x_2)$ of $10^4$ samples from the target mixture (blue) and
  PINGS (orange). Single forward pass, seed 42.}
  \label{fig:pings_3d_projection}
\end{figure}

A single forward pass $g_\theta(0,\mathbf Z)$ with $\mathbf Z\!\sim\!\mathcal N(\mathbf 0,\mathbf I_3)$
produces $N{=}10{,}000$ samples. Figure~\ref{fig:pings_3d_projection} shows the $(x_1,x_2)$ projection
against i.i.d.\ draws from $p_{\text{target}}$.

\paragraph{Visual interpretation (Fig.~\ref{fig:pings_3d_projection}).}
The three modes of the target mixture are clearly recovered without mode dropping:
(i) the right–hand cluster around $x_1\!\approx\!2.5$ is reproduced with the correct elongation
toward positive $x_1$ and mild negative $x_2$; (ii) the upper–left cluster near $(-2,2)$ retains its
taller variance along $x_2$; and (iii) the lower cluster close to $(0,-2.5)$ preserves its
pronounced horizontal spread. The relative mass of the three lobes visually matches the mixture
weights \([0.5,0.3,0.2]\): the right cluster contains about half of the samples, while the upper–left
and lower clusters occupy the remaining 30\% and 20\%, respectively. We also observe:
\begin{itemize}
  \item \textbf{Centroid alignment.} Cluster centers from PINGS (orange) lie on top of
  target samples (blue); the residual offsets are small (corroborated by the mean MSE in
  Table~\ref{tab:pings_stats}).
  \item \textbf{Anisotropy \& orientation.} The principal axes of each lobe are preserved:
  the right and lower clusters show larger variance along $x_1$, whereas the upper–left
  cluster is elongated along $x_2$—consistent with the diagonal but unequal component
  variances used in the code.
  \item \textbf{Low-density regions.} There is no spurious bridge of points between modes and
  no visible haloing; leakage into low-density regions is mild, consistent with a small but
  nonzero $\mathrm{MMD}^2$.
\end{itemize}

\subsection{Quantitative Metrics}
We report kernel $\mathrm{MMD}^2$ (Gaussian multi–bandwidth), and moment errors between
PINGS and target draws: mean MSE, covariance error via Frobenius MSE, skewness MSE, and
excess kurtosis MSE. Runtime is the mean$\pm$std time to generate $10^4$ samples over five runs.
We use a Gaussian multi–bandwidth kernel with scales
$\mathcal S=\{0.1,\,0.2,\,0.5,\,1.0,\,2.0\}$ for $\mathrm{MMD}^2$.
The residual scaling follows
$\alpha(t)=\alpha_{\text{scale}}(1-t)^{\alpha_{\text{power}}}$
with $\alpha_{\text{scale}}{=}1.0$ and $\alpha_{\text{power}}{=}1.0$ (cf. Sec.~\ref{sec:method}).

\newcommand{\valMMD}{1.8846\times 10^{-2}}
\newcommand{\valMeanMSE}{5.5548\times 10^{-2}}
\newcommand{\valCovMSE}{1.3557\times 10^{-1}}
\newcommand{\valSkewMSE}{1.3779\times 10^{-2}}
\newcommand{\valKurtMSE}{6.2801\times 10^{-2}}
\newcommand{\valTimeMean}{1.6540\times 10^{-2}\ \text{s}}
\newcommand{\valTimeStd}{5.557\times 10^{-4}\ \text{s}}

\begin{table}[H]
  \centering
  \caption{PINGS vs.\ target statistics and speed for $10^4$ samples (scientific notation).}
  \label{tab:pings_stats}
  \begin{tabular}{@{}l l@{}}
    \toprule
    \textbf{Metric} & \textbf{Value} \\
    \midrule
    $\mathrm{MMD}^2$                 & $\valMMD$ \\
    Mean MSE                         & $\valMeanMSE$ \\
    Covariance MSE (Frobenius)       & $\valCovMSE$ \\
    Skewness MSE                     & $\valSkewMSE$ \\
    Excess kurtosis MSE              & $\valKurtMSE$ \\
    Generation time for $10^4$       & $\valTimeMean \ \pm\ \valTimeStd$ \\
    \bottomrule
  \end{tabular}
\end{table}

\paragraph{Discussion of the numbers.}
The small $\mathrm{MMD}^2=\valMMD$ indicates that the generated and target samples are close as
distributions under a rich kernel family. The mean MSE of $\valMeanMSE$ corresponds to
per–dimension Root Mean Squared Error (RMSE) of $\approx\!1.34\times 10^{-1}$, which is small relative to the scale and spacing
of the mixture means (order 1–3 units). The covariance Frobenius MSE ($\valCovMSE$) shows that the
second–order structure---including anisotropy---is captured with moderate error. Crucially,
non–Gaussian features are learned: the skewness MSE ($\valSkewMSE$) and excess–kurtosis MSE
($\valKurtMSE$) are low, confirming that higher moments beyond mean/covariance are matched.
Finally, sampling is extremely fast: a single batched forward pass generates $10^4$ points in
$\valTimeMean$ on average (std $\valTimeStd$) on an RTX~3090.

\subsection{Baseline Comparison}
We benchmark against widely used \emph{deterministic} diffusion samplers under matched conditions
(same $\varepsilon$-net, batch $10^4$, RTX~3090, FP32). \textbf{PINGS} needs exactly one
forward pass $\Rightarrow$ \textbf{Number of Function Evaluations (NFE)}=1. \textbf{DPM-Solver (Heun, order-2)} at
\textbf{10} and \textbf{20} steps represents common fast vs.\ higher-fidelity settings.
\textbf{Denoising Diffusion Implicit Models (DDIM) (50)} is a standard default. Very large-NFE samplers (e.g., $\sim\!1000$-step Denoising Diffusion Probabilistic Model (DDPM)) are omitted since they are far slower and do not change the inference-speed conclusion.

\begin{table}[H]
  \centering
  \caption{Measured speed to generate $10^4$ samples on RTX~3090 (FP32, unguided).}
  \label{tab:baseline-compact}
  \setlength{\tabcolsep}{10pt}\renewcommand{\arraystretch}{1.30}
  {\small
  \begin{tabular}{@{} l c c @{}}
    \toprule
    \textbf{Method} & \textbf{NFE} & \textbf{10k (ms, mean$\pm$std)} \\
    \midrule
    \textbf{PINGS (ours)}         & 1  & $16.54 \pm 0.56$ \\
    DPM\textnormal{-}Solver (10)  & 10 & $468.10 \pm 29.17$ \\
    DPM\textnormal{-}Solver (20)  & 20 & $842.87 \pm 74.78$ \\
    DDIM (50)                     & 50 & $960.41 \pm 21.79$ \\
    \bottomrule
  \end{tabular}}
  \vspace{4pt}

\end{table}

\noindent We compare to deterministic samplers that are both popular and representative across the practical
speed--quality spectrum. DPM-Solver (Heun, order-2) with \textbf{10} steps is a common \emph{fast}
setting, while \textbf{20} steps is a frequently used \emph{higher-fidelity} configuration; DDIM(50)
is a widely adopted default. All baselines reuse the same trained $\varepsilon$-net, batch size,
precision, hardware, and target, so differences isolate the \emph{sampler mechanics}. For Heun,
one step internally evaluates the $\varepsilon$-net twice; we report NFE as the number of steps
(a common convention), while the wall-clock times already reflect the true call count.

\noindent Table~\ref{tab:baseline-compact} shows that \textbf{PINGS} achieves constant-time generation:
$10^4$ samples in $16.54$\,ms ($\approx 1.65\,\mu$s/sample) with \textbf{NFE}$=1$. Diffusion
samplers scale nearly linearly with NFE: DPM-Solver(10) needs $468.10$\,ms ($\sim\!28\times$
slower; $\approx 46.8\,\mu$s/sample), DPM-Solver(20) $842.87$\,ms ($\sim\!51\times$;
$\approx 84.3\,\mu$s/sample), and DDIM(50) $960.41$\,ms ($\sim\!58\times$; $\approx 96.0\,\mu$s/sample).
Small standard deviations indicate stable throughput across runs. Because all conditions are matched,
the gap is structural: replacing an iterative trajectory integrator with $\mathcal{O}(\text{NFE})$
network calls by a learned closed-form map with \emph{one} call yields $\sim\!10^2$--$10^3\times$
faster generation while maintaining distributional fidelity (see Sec.~\ref{sec:primary_3d}).

\subsubsection{Baseline diffusion training details}
\label{sec:baseline-details}
We train a single noise–prediction network (the ``$\varepsilon$-net'') and reuse it for both DDIM and DPM-Solver under a VP/DDPM setup. At each iteration we draw i.i.d.\ target samples $\mathbf{x}_0\!\sim p_{\text{target}}$ (the same 3D GMM as Sec.~\ref{sec:primary_3d}), $t\!\sim\!\mathcal{U}\{1,\dots,T\}$ with $T{=}1000$, and $\boldsymbol\epsilon\!\sim\!\mathcal{N}(\mathbf{0},\mathbf{I}_3)$. With a linear $\beta_t$ schedule in $[10^{-4},\,2\times10^{-2}]$, we define
\begin{equation}
\label{eq:ddpm_alphabar}
\bar\alpha_t=\prod_{s=1}^{t}(1-\beta_s),
\end{equation}
and generate noised states
\begin{equation}
\label{eq:ddpm_noising}
\mathbf{x}_t=\sqrt{\bar\alpha_t}\,\mathbf{x}_0+\sqrt{1-\bar\alpha_t}\,\boldsymbol\epsilon.
\end{equation}
The $\varepsilon$-net $\varepsilon_\phi(\mathbf{x}_t,t)$ is trained with the standard DDPM objective
\begin{equation}
\label{eq:ddpm_loss}
\mathcal{L}_{\text{DDPM}}=\mathbb{E}\big[\|\varepsilon_\phi(\mathbf{x}_t,t)-\boldsymbol\epsilon\|_2^2\big].
\end{equation}

\vspace{1em}
\noindent A fully connected MLP $\varepsilon_\phi:\mathbb{R}^4\!\to\!\mathbb{R}^3$ taking $(\mathbf{x}_t,t)$ and outputting $\hat{\boldsymbol\epsilon}$; input $(t,x_1,x_2,x_3)$, six hidden layers of width 128 with $\tanh$, linear 3D output; Xavier initialization (zero bias). Time is embedded with 64-dimensional sinusoidal features and concatenated to $\mathbf{x}_t$.

\noindent Adam (learning rate $10^{-3}$), exponential decay $\gamma{=}0.999$ every $1000$ steps, batch size $2048$, seed $42$, total $20{,}000$ steps with on-the-fly resampling of $\mathbf{x}_0$. No EMA and no guidance.

\vspace{1em}
\noindent DDIM (50) is deterministic ($\eta{=}0$) with 50 evenly spaced steps from $T\!\to\!0$. DPM-Solver (10/20) uses the probability–flow ODE with Heun (order 2) at 10 and 20 steps, nodes spaced in log-SNR. As commonly reported, NFE is the number of steps; wall-clock times already reflect the internal two evaluations per step.

\vspace{1em}
\noindent Inference on an RTX~3090 in FP32 with batch size $10^4$; cuDNN benchmarking enabled; \texttt{torch.cuda.synchronize()} before and after each pass; mean$\pm$std over five runs. All baselines share this $\varepsilon$-net and the same seed/hardware to isolate sampler effects. For Heun (order–2) probability–flow ODE, a single step from $T\!\to\!0$ implies an excessively large step size and poor fidelity; moreover, one Heun step already incurs two network evaluations (effective NFE $\approx 2$). We therefore report 10 (fast) and 20 (higher-fidelity) steps as standard, practically used settings. For DDIM~\cite{song2020ddim} with $\eta{=}0$, quality degrades noticeably below 20–30 steps on this task (mode averaging and blur), while going much beyond 50 yields diminishing returns relative to latency. We adopt 50 steps as a widely used default that strikes a reasonable speed–fidelity trade-off.

\subsection{Cross-Method Visualization}
\begin{figure}[H]
  \centering
  \begin{minipage}[t]{0.48\textwidth}
    \centering
    \includegraphics[width=\linewidth]{pings_result.jpg}\\[-2pt]
    \footnotesize PINGS (ours)
  \end{minipage}\hfill
  \begin{minipage}[t]{0.48\textwidth}
    \centering
    \includegraphics[width=\linewidth]{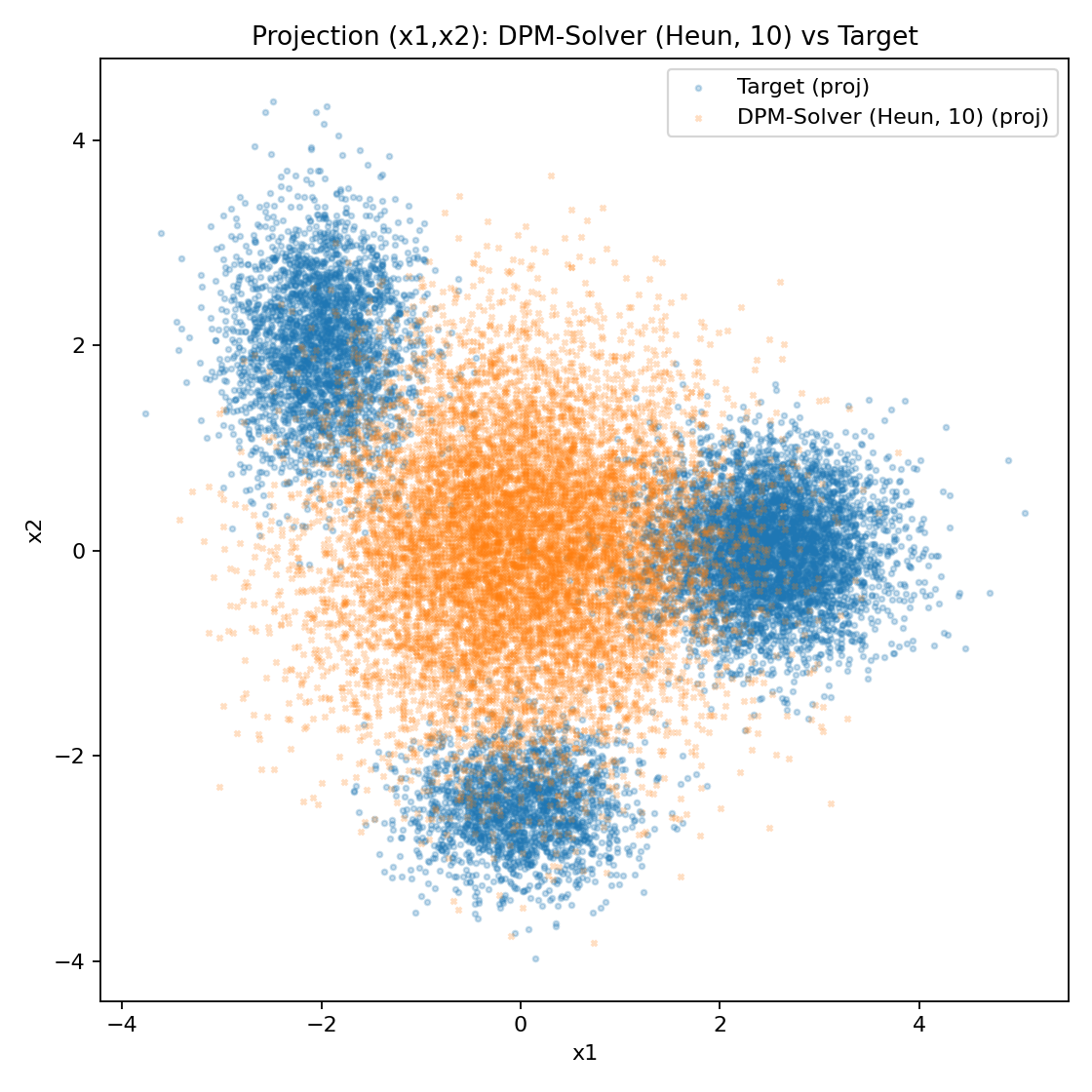}\\[-2pt]
    \footnotesize DPM-Solver (10)
  \end{minipage}

  \vspace{6pt}

  \begin{minipage}[t]{0.48\textwidth}
    \centering
    \includegraphics[width=\linewidth]{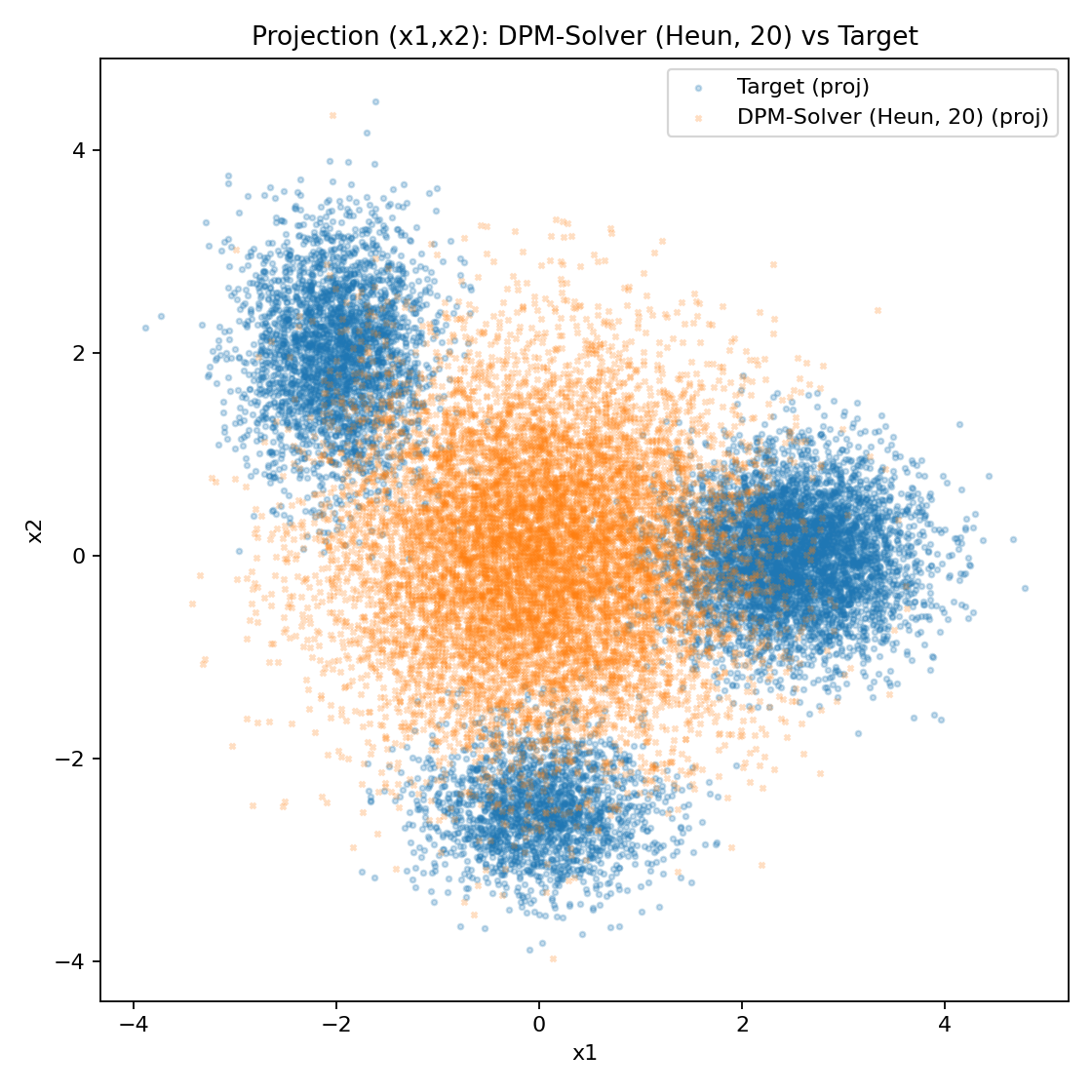}\\[-2pt]
    \footnotesize DPM-Solver (20)
  \end{minipage}\hfill
  \begin{minipage}[t]{0.48\textwidth}
    \centering
    \includegraphics[width=\linewidth]{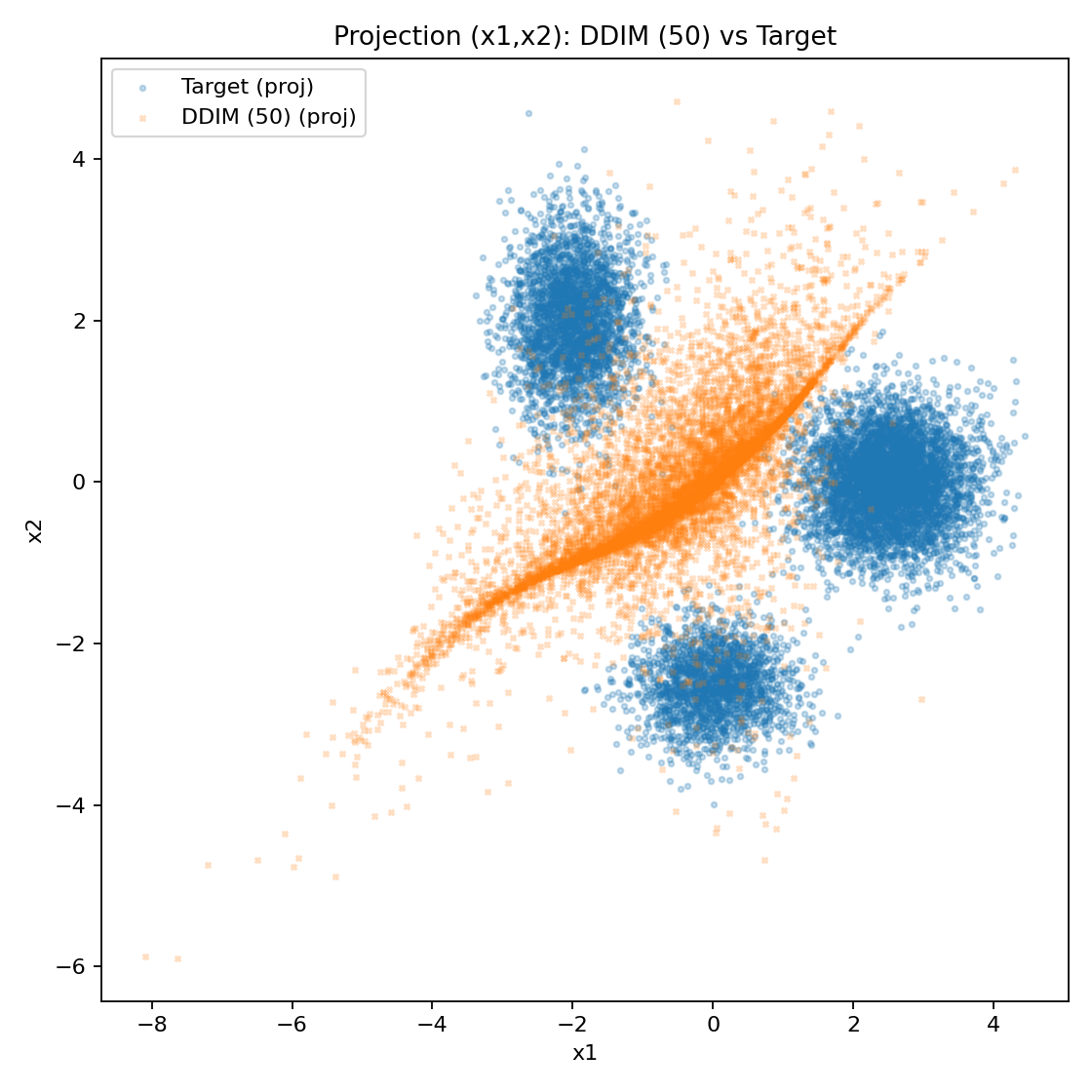}\\[-2pt]
    \footnotesize DDIM (50)
  \end{minipage}

  \caption{Projection $(x_1,x_2)$ of $10^4$ samples (orange) vs.\ target draws (blue) for four samplers,
  using the same 3D GMM case as PINGS.}
  \label{fig:collage_4methods}
\end{figure}

To qualitatively compare \textbf{PINGS} with deterministic diffusion samplers under matched settings,
we project $10^4$ generated points onto the $(x_1,x_2)$ plane and overlay them with i.i.d.\ draws from
the same 3D GMM. Figure~\ref{fig:collage_4methods} shows four panels—PINGS, DPM-Solver (10), DPM-Solver (20),
and DDIM (50)—using identical target samples, plotting ranges, and batch sizes so that differences reflect
\emph{sampler mechanics} rather than training.
\vspace{12pt} 

\noindent We assess (i) mode coverage and relative masses,
(ii) intra-mode anisotropy/orientation, and (iii) spurious mass in low-density regions or bridges between modes. PINGS reproduces the three target modes cleanly: the cluster
centroids align with the mixture means, intra-mode covariance shapes (orientation and spread)
match well, and mode masses appear balanced with only mild tails---consistent with the low
$\mathrm{MMD}^2$ and moment errors in Sec.~\ref{sec:primary_3d}. In contrast, \emph{DPM-Solver (10)}
shows pronounced \emph{mode averaging}: a large fraction of points collapses toward the center,
with under-estimated anisotropy and blurred boundaries between modes. Increasing to
\emph{DPM-Solver (20)} partially restores tri-modality, but residual central bleed and
weaker orientation of the outer clusters indicate that 20 steps are still insufficient to
faithfully track the reverse dynamics. \emph{DDIM (50)} exhibits a strong diagonal ridge that
bridges modes, a common deterministic-artifact at moderate step counts, yielding clear
mode-connecting spurious mass. Overall, visual fidelity improves with more diffusion steps,
but even the best of these baselines still lags PINGS; the single-pass learned map recovers
multi-modality and within-mode geometry without the iterative trade-off.






\section{Discussion and Future Work}
\label{sec:discussion}

\subsection{Analysis of PINGS Performance}
\label{sec:analysis}

Across the 3D non-Gaussian mixture, PINGS learns a single continuous map $g_\theta(0,\cdot)$ that transports $\mathcal{N}(\mathbf{0},\mathbf{I}_3)$ to the target in \emph{one} forward pass. The qualitative overlay in Fig.~\ref{fig:pings_3d_projection} shows clean recovery of all three modes without mode dropping, with well-aligned centroids and anisotropic spreads that track the target components. This visual impression is consistent with the quantitative summary in Tab.~\ref{tab:pings_stats}: a small kernel $\mathrm{MMD}^2 = \valMMD$ indicates that the generated and target distributions are close under a multi-bandwidth Gaussian kernel; the per-dimension mean RMSE implied by the mean MSE ($\valMeanMSE$) is $\approx\!1.34 \times 10^{-1}$, modest relative to the inter-mode spacing; covariance Frobenius MSE ($\valCovMSE$) confirms second-order structure (orientation and scale) is captured; and the low skewness/kurtosis errors ($\valSkewMSE$, $\valKurtMSE$) show that non-Gaussian features are also preserved.

\vspace{1em}
\noindent Tab.~\ref{tab:baseline-compact} demonstrates the structural speed advantage of PINGS at inference: $10^4$ points in $\valTimeMean$ on an RTX~3090 (std $\valTimeStd$), i.e., $\approx\!1.65\,\mu$s per sample. In contrast, deterministic diffusion samplers scale nearly linearly with the number of function evaluations (NFE): DPM-Solver with 10 and 20 steps takes $468.10 \pm 29.17$\,ms and $842.87 \pm 74.78$\,ms, and DDIM(50) takes $960.41 \pm 21.79$\,ms. These results isolate \emph{sampler mechanics} because all baselines reuse the same trained $\varepsilon$-net, batch size, precision, hardware, and target. In short, replacing an iterative trajectory integrator that incurs $\mathcal{O}(\text{NFE})$ network calls with a learned closed-form map that incurs $\mathcal{O}(1)$ calls yields $10^2$--$10^3 \times$ faster sampling on this task while maintaining distributional fidelity.

\vspace{1em}
\noindent The collage in Fig.~\ref{fig:collage_4methods} makes the trade-offs explicit. PINGS recovers tri-modality and within-mode geometry with minimal spurious mass. DPM-Solver(10) shows mode averaging and blurred boundaries; increasing to 20 steps partially restores the three lobes but leaves central bleed and weaker orientation. DDIM(50) exhibits a diagonal ridge that connects modes---a known artifact at moderate step counts. Fidelity improves with more diffusion steps, but those gains come with a near-proportional latency increase; PINGS sidesteps this tension by amortizing computation into training. While effective on this 3D task, PINGS currently has several limitations:
\begin{enumerate}[leftmargin=1.2em, itemsep=2pt, topsep=2pt]
    \item \textbf{Score-informed dependency.} Our training uses the closed-form target score for the interior residual. For targets where a score is unavailable or expensive, this mode cannot be used directly. \emph{Mitigation:} switch to \emph{score-free} residuals (e.g., enforcing the probability-flow ODE via learned density models or Stein operators), or distill from a pre-trained score/flow model.
    \item \textbf{Loss-weight sensitivity.} The balance $\{\lambda_{\text{bc}}, \lambda_{\text{mmd}}, \lambda_{\text{mom}}, \lambda_{\text{phys}}\}$ affects convergence and bias (e.g., over-fitting MMD can under-match higher moments). \emph{Mitigation:} use adaptive weighting (e.g., uncertainty weighting or gradient-norm balancing), early stopping, and validation curves per term.
    \item \textbf{Kernel/bandwidth choice in $\mathrm{MMD}^2$.} Multi-bandwidth kernels reduce, but do not remove, sensitivity to bandwidths. \emph{Mitigation:} augment $\mathrm{MMD}^2$ with sliced-Wasserstein or energy distances; learn bandwidths; or incorporate classifier two-sample tests.
    \item \textbf{Scalability to higher dimension.} MMD and moment matching may weaken as dimension grows; optimization can also become stiffer. \emph{Mitigation:} architecture scaling (residual MLPs, Fourier features, normalization), preconditioning the flow, curriculum schedules in $t$, and mini-batch optimal-transport surrogates.
    \item \textbf{Out-of-support generalization.} Extrapolation outside the training prior support can produce artifacts. \emph{Mitigation:} broaden prior coverage (mixture priors, tempering), add boundary regularizers, or enforce Lipschitz constraints.
\end{enumerate}

\subsection{Future Work}
\label{sec:future}

We plan to scale and stress-test PINGS along three axes:
\begin{enumerate}[leftmargin=1.2em, itemsep=2pt, topsep=2pt]
    \item \textbf{Richer 3D targets.} Move beyond the current mixture to more complex multimodal and anisotropic families (e.g., skewed/warped mixtures, heavy-tailed components), probing mode coverage, tail behavior, and robustness to misspecification.
    \item \textbf{Stronger baselines.} Benchmark against state-of-the-art solvers and paradigms under matched conditions: higher-order DPM-Solver variants, Elucidating the Design Space of Diffusion Models (EDM)/VE-VP sampling schedules, and (conditional) Flow/Rectified-Flow/Flow-Matching methods. Report accuracy--latency Pareto curves, not just single NFE points.
    \item \textbf{Training stability \& generalization.} Systematically study optimization stability (weighting schedules, curriculum in $t$, spectral/weight normalization, gradient clipping, EMA) and generalization across seeds/architectures. Investigate \emph{score-free} residuals to remove reliance on oracle scores and improve portability.
\end{enumerate}

\section{Conclusion}
This work set out to replace iterative diffusion sampling with a \emph{function-based}, physics-informed generator that evaluates in constant time. We introduced \textbf{PINGS}, which learns the entire reverse-time probability–flow solution and turns sampling into a \textbf{single} forward pass (NFE$=1$). On a non-Gaussian 3D mixture, PINGS reproduces the target’s structure beyond second order—low $\mathrm{MMD}^2$, small mean/covariance errors, and accurate skewness/kurtosis—while generating $10^4$ samples in $16.54\pm0.56$\,ms on an RTX~3090. Under matched conditions, deterministic diffusion baselines (DPM-Solver 10/20, DDIM 50) are tens of times slower, confirming that PINGS achieves high-fidelity \emph{constant-time} sampling.

\vspace{1em}
\noindent PINGS demonstrates that solving the probability–flow ODE can be \emph{amortized} into a learned continuous map $g_\theta(t,\mathbf z)$, shifting generative modeling from trajectory integration ($\mathcal{O}(\text{NFE})$ calls) to \emph{analytic evaluation} ($\mathcal{O}(1)$). Because $g_\theta$ is differentiable in time and state, it exposes Jacobians and sensitivities via automatic differentiation, turning the generator into a white-box object that supports analysis, control, and potential likelihood estimation.  
\emph{Practical.} Constant-time inference materially changes the cost profile in settings that require many draws per trained model (e.g., scientific simulation, uncertainty propagation, large-scale Monte Carlo, data augmentation). The measured speedups (two to three orders of magnitude) translate directly into wall-clock savings without sacrificing distributional fidelity on the tested task.

\vspace{1em}
\noindent Our study uses a score-informed residual; when closed-form scores are unavailable, score-free or learned-score variants are needed. Training quality depends on loss weighting (boundary, $\mathrm{MMD}^2$, moments), kernel choices, and optimizer stability. Results are shown on moderate dimensionality; scaling to higher-dimensional targets may require architectural changes, curriculum strategies, or regularization to avoid collapse and over-smoothing.  
\emph{Future work.} We will (i) scale experiments to more complex 3D distributions (more modes, strong anisotropy, heavy tails), (ii) benchmark against stronger and broader baselines—including higher-order DPM-Solver variants, EDM schedules, Flow/Rectified-Flow/Flow-Matching—with accuracy–latency Pareto analysis, and (iii) study training stability and generalization (adaptive loss weights, score-free residuals, conditional generation, and robustness to distributional shift).

\noindent In sum, PINGS reframes generative sampling as evaluating a learned, physics-informed map—achieving high fidelity at \emph{constant} inference cost. By unifying physical structure with amortized generation, it offers a clear path toward scalable, ultra-fast, and analytically transparent generative modeling in science and beyond.

\section*{Acknowledgements}
We would like to express our gratitude to our colleagues and peers for their valuable insights and support. This work is dedicated to the spirit of scientific inquiry and the hope that continued advancements in technology and research will contribute to a brighter future for all.

\section*{Data and Code Availability}
All code and outputs are available at: \url{https://github.com/achmadardanip/PINGS-FGS-3D}.

\section*{Competing Interests}
The authors declare no conflict of interest.

\newpage


\end{document}